\documentclass{article} 
\usepackage{iclr2025_conference,times}

\usepackage{amsmath,amsfonts,bm}

\def\eqref#1{equation~\ref{#1}}

\def\1{\bm{1}}

\DeclareMathAlphabet{\mathsfit}{\encodingdefault}{\sfdefault}{m}{sl}
\SetMathAlphabet{\mathsfit}{bold}{\encodingdefault}{\sfdefault}{bx}{n}

\usepackage{hyperref}
\usepackage{url}
\usepackage{subcaption}

\usepackage{graphicx}
\usepackage{xcolor}

\usepackage{amsmath}
\usepackage{enumitem}

\usepackage{algorithm}
\usepackage{algpseudocode}

\usepackage{sidecap}
\usepackage{array}
\usepackage{ulem} 
\usepackage[utf8]{inputenc} 
\usepackage[T1]{fontenc}    
\usepackage{hyperref}       
\usepackage{url}            
\usepackage{booktabs}       
\usepackage{amsfonts}       
\usepackage{nicefrac}       
\usepackage{microtype}      
\usepackage{xcolor}         

\usepackage{ulem}  

\usepackage{graphicx}

\usepackage{fp}

\newcommand{\roundto}[2]{%
    \FPsub\tempresult{#1}{1}%
    \FPmul\tempresult{\tempresult}{100}%
    \FPround\result{\tempresult}{1}%
    \result\%
}

\title{Decoupled Relative Learning Rate Schedules}

\author{%
  Jan Ludziejewski \\
  University of Warsaw \\
  IDEAS NCBR 
  \And Jan Małaśnicki \\
  University of Warsaw \\
  IDEAS NCBR \\
  \And Maciej Pióro \\
  Polish Academy of Sciences \\
  IDEAS NCBR \\
  \And Michał Krutul \\
  University of Warsaw \\
  IDEAS NCBR \\
  \And Kamil Ciebiera \\
  University of Warsaw \\
  IDEAS NCBR \\
  \And Maciej Stefaniak \\
  University of Warsaw \\
  IDEAS NCBR \\
  \And Jakub Krajewski \\
  University of Warsaw \\
  IDEAS NCBR \\
  \And Piotr Sankowski \\
  University of Warsaw \\
  MIM Solutions \\
  \And Marek Cygan \\
  University of Warsaw \\
  Nomagic \\
  \And Kamil Adamczewski \\
  Wroclaw Tech\\
  IDEAS NCBR \\
  \And Sebastian Jaszczur \\
  University of Warsaw \\
  IDEAS NCBR \\
}

\newcommand{\moe}[1]{$\text{MoE}_{8\times#1\text{M}}$}
\newcommand{\dense}[1]{$\text{Dense}_{#1\text{M}}$}

\iclrfinalcopy 
\begin{document}

\maketitle

\begin{abstract}
In this work, we introduce a novel  approach for optimizing LLM training by adjusting learning rates across weights of different components in Transformer models. Traditional methods often apply a uniform learning rate across all network layers, potentially overlooking the unique dynamics of each part. Remarkably, our introduced relative learning rates, RLRS, method accelerates the training process by up to {$23\%$}, particularly in complex models such as Mixture of Experts (MoE). Hyperparameters of RLRS can be efficiently tuned on smaller models and then effectively reused on models up to $27\times$ larger. This simple and effective method results in a substantial reduction in training time and computational resources, offering a practical and scalable solution for optimizing large-scale neural networks.

\noindent\let\thefootnote\relax\footnote{Correspondence to jahulas@gmail.com, jan.malasnicki@gmail.com, kamil.adamczewski@pwr.edu.pl}
\end{abstract}

\section{Introduction}

The learning rate is a crucial hyperparameter in Deep Learning, determining the size of the steps that the optimization algorithm takes when updating model parameters during training. In the context of Transformers, widely used for tasks in Natural Language Processing (NLP) and other areas, the learning rate significantly impacts the model's convergence and overall performance. While higher learning rates, with larger updates to the model, may generally converge faster, they can also introduce instabilities. Therefore, the learning rate must be carefully chosen to balance the speed and stability of the training process.

\begin{figure}[h]
    \centering
    \includegraphics[width=\linewidth]{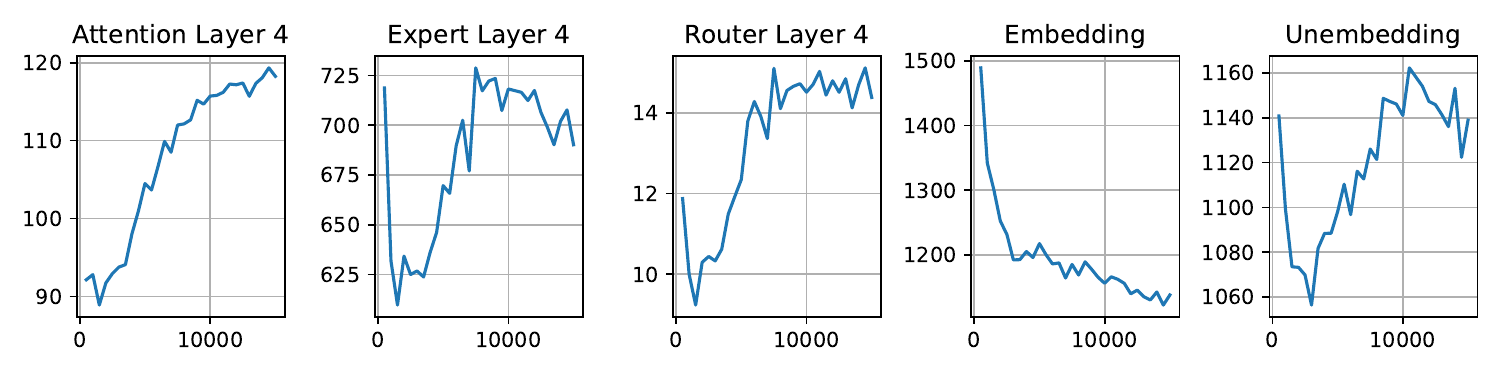}
    
                    
    
    \vspace{0.2cm} 
    \caption{Magnitudes of weight updates for different components of the Transformer with MoE (trained without RLRS), right before applying learning rate.} 
    \label{fig:updates}
\end{figure}

At the same time, modern Deep Learning architectures are not homogeneous, with different parts having distinct structures, serving varied purposes, and exhibiting unique behaviors. Importantly, they also have individual training dynamics. As seen in Figure~\ref{fig:updates}, weight updates for different model parts follow different patterns during training. This variability can result in components behaving differently depending on the training phase, which may be problematic in some cases. For example, in Mixture of Experts (MoE) models, the Router often stabilizes early in training, leading to deterministic routing to the Experts~\citep{xue2024openmoe}. 



Given the diversity of layers within a model, it is reasonable to expect that their requirements would vary, particularly when balancing training speed and stability. Despite this, a uniform learning rate is often applied across all modules. A common practice, for example, is to reduce the learning rate for the whole model after the introduction of an MoE layer due to instabilities~\citep{pmlr-v162-rajbhandari22a}. As a result, hyperparameters are typically tuned for the entire network, even if the instabilities may originate from a single layer. In this work, we challenge the implicit assumption of a global learning rate.
Motivated by the heterogeneity of Transformer's modules, we ask ourselves: can we improve the training procedure by tailoring the learning rate schedules to different model's components? 

To answer this question, we decouple learning rates in Transformers and tune them individually for various model components—including Embedding, Unembedding, Attention, Feed-Forward, and, in the case of Mixture of Experts architecture, Router and Experts—we enhance the model's overall performance and stability by tailoring the learning rate to meet the specific needs of each component. 

Commonly used adaptive optimizers, like AdamW \citep{loshchilov2019decoupled}, show the importance of well-adjusted learning rates during training. However, adaptive optimizers generally treat all layers in the same way and do not differentiate between them. Our proposed RLRS is used on top of an adaptive optimizer (AdamW in this work, but RLRS is independent of the optimizer), and we show that adjusting learning rate schedules for individual model components brings improvements.

Furthermore, we propose a straightforward scheme to adjust relative learning rate values that can be effectively scaled to models larger by orders of magnitude. This approach eliminates the need for extensive hyperparameter searches for larger models, resulting in significant computational savings and enhancing its practical applicability. (This could also be combined with Tensor Programs \citep{yang2022tensor}, see discussion in Section~\ref{future:tensor_programs}.)

In essence, we propose the following approach: first, relative learning rates, RLRS, should be tuned on a small model; later, the same RLRS can be reused when training the model's significantly larger counterpart.
Our method is easy to implement, with no additional overhead required, apart from the relatively inexpensive hyperparameter search on the small model. While tailored to our specific training setup, the relative LR values configuration shared in this work have proven robust across a range of models sizes and training durations, making them an excellent starting point. 
Additionally, we provide an analysis showing how these values, obtained using automated methods, align with our intuitive understanding of Transformer training. In summary,

\begin{itemize}
    \item We propose distinct, relative learning rate schedules (RLRS) tailored for different components of a Transformer model, optimizing each part individually for better overall performance.
    \item We demonstrate performance improvements of the introduced method in standard Transformers, and achieving  even larger speedup in the case of Mixture of Experts (MoE) based models, highlighting the importance of relative learning rates for more complex models. 
    \item We demonstrate that the hyperparameters tuned on small models extrapolate to larger models, showing that our approach generalizes effectively across different architecture sizes. 
    
\end{itemize}

    
                
    

\section{Decoupled Relative Learning Rate Schedules}
 \begin{figure}
    \centering
    \includegraphics[width=\linewidth]{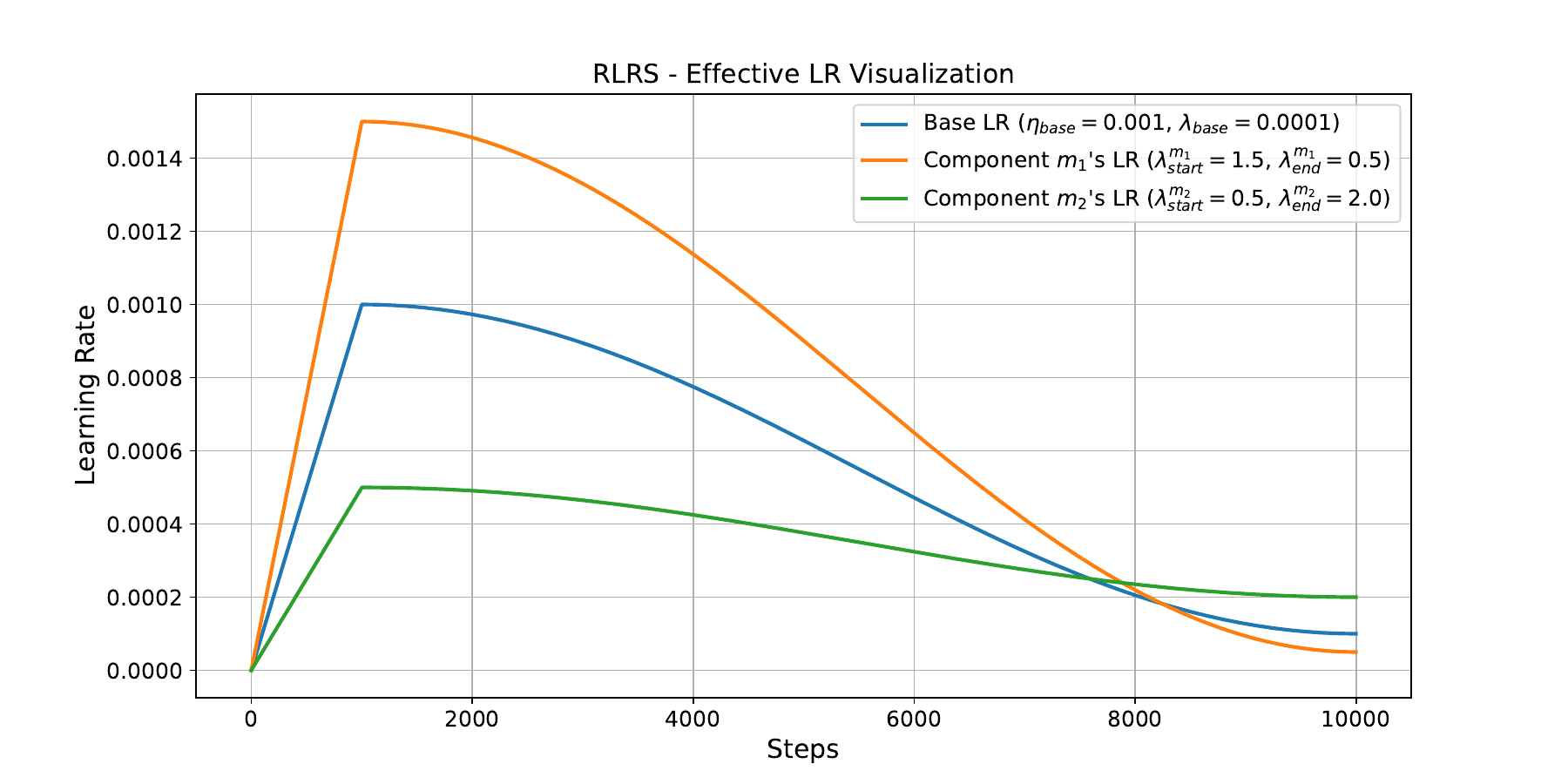}
    \label{fig:rlrs_vis}
    \caption{Visualization of distinct cosine learning rate schedules for each model component, compared against the base model’s learning rate.}
\end{figure}

We define a \textit{decoupled learning rate} as a separate learning rate schedule for different layer types (also called parts, modules, or components). Decoupled learning rate schedules enable the learning procedure to focus on different components during various phases of a model's pretraining, facilitating a more targeted and efficient optimization process.

We specify decoupled learning rates following the structure of the cosine learning rate scheduler~\citep{loshchilov2016sgdr}, widely used for training Large Language Models (LLMs)~\citep{touvron2023llama, hoffmann2022training}. The cosine scheduler adjusts the learning rate over time according to a cosine function, starting with a high learning rate that gradually decays to a minimum value in a smooth, nonlinear manner.


We define the base learning rates, shared across modules, using parameters:

\begin{itemize}[topsep=0pt, leftmargin=*] 
\item \textit{Base LR} ($\eta_{\text{base}}$) --- the reference learning rate for the entire model. In a typical cosine schedule, it is the peak learning rate.

\item \textit{Base LR Final Fraction} ($\alpha_{\text{end}}$) --- the reference fraction of the base learning rate at the end of the training. In a typical cosine schedule, the final learning rate $\eta_{\text{end}} = \eta_{\text{base}} \times \alpha_{\text{end}}$. 


 \end{itemize}


The cosine scheduler adjusts the learning rate following a cosine curve over a specified number of iterations.  The learning rate $\eta_t$ at step $t$ is computed using the cosine function: $\eta_t = \eta_{\text{end}} + \frac{1}{2} (\eta_{\text{start}} - \eta_{\text{end}}) \left(1 + \cos\left(\frac{t}{T} \pi\right)\right)$, where $t$ is the current step, and $T$ is the total number of steps.

For each component $m$ of the model, we further define a learning rate scaling factors \textit{relative} to the base learning rates defined above:

\begin{itemize} [topsep=0pt, leftmargin=*]
\item \textit{Relative Start LR} ($\lambda_{\text{start}}^m$) --- the scaling factor of the base learning rate at the beginning of training. 

\item \textit{Relative End LR} ($\lambda_{\text{end}}^m$) --- the scaling factor of the final learning rate at the end of training. 
\end{itemize}

Thus, the \textit{decoupled learning rates} $\eta_{\text{start}}^m$ and $\eta_{\text{end}}^m$ for a component $m$ are defined as:



\begin{align}
\eta_{\text{start}}^m &= \eta_{\text{base}} \times \lambda_{\text{start}}^m \\
\eta_{\text{end}}^m &= \eta_{\text{base}} \times \alpha_{\text{end}} \times \lambda_{\text{end}}^m
\end{align}

These values are adjusted for each Transformer component. In this work, we distinguish the following layer modules $m$: Embedding, Attention, Unembedding, and additionally, for dense models, the Feed-Forward layer, and for Mixture of Experts models, the Expert and Router layers.


Thus, for a given model, one needs to obtain the baseline learning rate $\eta_{\text{base}}$ and the relative learning rates, $\lambda_{\text{start}}^m$, $\lambda_{\text{end}}^m$. In Section~\ref{app:local_search}, we show how we find them in practice. However, in the next section, we demonstrate that we do not need to tune the relative learning rates for every model separately, as the same set of values remains robust across a range of model sizes. 




\subsection{Preserving Relative Values}


Directly tuning relative learning rate values on large models may be impractical due to significant computational costs. To address this, we propose a method that fine-tunes these values on a smaller proxy model and then transfers them to a larger model. This approach significantly reduces the need for costly tuning on large models, offering substantial computational savings.

Our method, described in Algorithm~\ref{algo}, involves conducting a search for optimal values on smaller models under the assumption that these relative values extrapolate effectively to larger models. This search consumes only a fraction of the training time required for large models.

\begin{algorithm}
\caption{Relative Learning Rate Adjustment Procedure}
\begin{algorithmic}[1]
\State Determine $\eta_{\text{base}}$ for a small model.
\State For each module $m$, find relative values $\lambda_{\text{start}}^m$ and $\lambda_{\text{end}}^m$ on a small model.
\State Determine base learning rate $\eta_{\text{base}}$ for the large model.
\State Apply relative learning rates $\lambda_{\text{start}}^m$ and $\lambda_{\text{end}}^m$ from the small model to the large model.
\end{algorithmic}
\label{algo}
\end{algorithm}

While we do not claim that $\lambda_{\textit{start}}^m$ and $\lambda_{\textit{end}}^m$ values are optimal for larger models, they are straightforward to use and yield substantial improvements, as shown in the next section. We leave the investigation of optimal extrapolation as future work. 


Algorithm~\ref{algo} presents a simple way to introduce relative learning rates independently of the base learning rate and to apply them to both small and large models. Moreover, as we see in Table \ref{tab:extrapol}, the base learning rate $\eta_{\text{base}}$ tuned with relative rates rarely differs from the optimal learning rate for the baseline non-RLRS model with the same configuration. Therefore, if an already tuned learning rate for a large model is available, we can reuse it as our base learning rate $\eta_{\text{base}}$ with relative schedules. Then further applying relative rates would yield substantial improvement without additional tuning on a large scale, which can be useful in practice.

%



We provide the details of our implementation of the hyperparameter search in Section~\ref{app:local_search} and Algorithm~\ref{algo_local}. Moreover, additional methods for adjusting the $\eta_{\text{base}}$ on extrapolation are considered in Section~\ref{future:tensor_programs}.


\section{Results}

We first provide a detailed setup of our experiments. Then, we present details on how to find relative learning rates, and showcase the improvements obtained via the proposed decoupled relative learning rate schedules. Subsequently, we provide some interpretation of the presented relative rates. For reproducibility purposes, we will provide the complete code and configuration files used in our experiments in a public repository soon.

\subsection{Experimental Setup}




All models in this study are decoder-only Transformers trained on the C4 dataset~\citep{raffel2023exploring}. We use the GPT-2 tokenizer~\citep{radford2018improving} and optimize with AdamW~\citep{loshchilov2019decoupled}. Training follows a cosine decay schedule with linear warmup for the first $1\%$ of steps. For stability, weights are initialized with a truncated normal distribution at a reduced scale, as suggested by \citet{fedus2022switch}. Mixed precision training is applied, with Attention and Router components computed at high precision. The models employ SwiGLU activation and Token Choice routing with $8$ Experts, $1$ of which is activated per token. Two auxiliary losses are used for the Router: a z-loss weighted at $0.001$~\citep{zoph2022st} and load balancing weighted at $0.01$~\citep{fedus2022switch}. Compute-optimal training durations align with \citet{hoffmann2022training}, calculated for MoE as $20$ times the number of active parameters, excluding Embedding and Unembedding, as per \citet{ludziejewskiscaling}. We also report results on an overtrained \moe{113} model with a token-to-active-parameter ratio nearing $130$. For extrapolations, we fine-tune base learning rates for both relative learning rates, RLRS, and the baseline on a grid of {$1e{-n}$, $2e{-n}$, $5e{-n}$}. By tuning learning rates separately, we ensure that performance differences arise from the effects of relative values rather than from base learning rates.
For both dense and MoE models, the weight decay value has been optimized to $0.1$, the initialization scale to $0.15$, and \textit{Base LR Final Fraction} ($\lambda_{\text{base}}$) to $0.04$ for MoE and $0.06$ for dense. 

\begin{table}[H]
    \centering

\begin{tabular}{cccccccc}
\toprule
Type & Active Params & Total Params & $d_{\textit{model}}$ & $n_{\textit{layers}}$ & $n_{\textit{experts}}$ & \texttt{BS} & \texttt{SL} \\
\midrule
\dense{34} & $33.6$M & $33.6$M & $512$ & $8$ & $ $ & $256$ & $512$ \\
\moe{34} & $33.6$M   & $210$M  & $512$ & $8$ & $8$ & $256$ & $512$ \\
\midrule
\dense{113} & $113$M & $113$M  & $768$ & $12$ & $ $ & $256$ & $512$ \\
\moe{113} & $113$M   & $708$M  & $768$ & $12$ & $8$ & $256$ & $512$ \\
\midrule
\dense{906} & $906$M & $906$M  & $1536$ & $24$ & $ $ & $384$ & $1024$ \\
\moe{906} & $906$M   & $5.67$B & $1536$ & $24$ & $8$ & $384$ & $1024$ \\
\bottomrule
\end{tabular}
\vspace{0.1cm}
    \caption{Models used in this paper. \texttt{BS} indicates batch size, and \texttt{SL} indicates sequence length. Active parameters exclude Embedding and Unembedding parameters.}
    \label{tab:my_label}
\end{table}

\paragraph{Speed-up metric.} In Tables~\ref{tab:small} and~\ref{tab:extrapol}, we present a speedup metric that measures how much faster a training process becomes when relative rates are applied. It is calculated using $(\frac{T_{\text{base}}}{T_{\text{relative}}} - 1) \times 100\%$, where $T_{\text{base}}$ is the number of steps performed in the standard training with a base learning rate, and $T_{\text{relative}}$ is the number of steps incurred until the loss of the training with the relative learning rate schedule exceeds the baseline loss. It is important to note that using this metric likely underestimates the improvement of our method since for relative learning rate training steps, when we compute the speedup, the cosine schedule has not yet reached its end.
We perform three runs for each configuration, and for each of them, we measure the loss per $1\%$ of all training steps . The speedup is then calculated over the means of $3$ runs. To reduce variance from random data seeds, we use $3$ specified data seeds for each model type comparison.

\paragraph{Local search.} Our approach involves optimizing a set of relative learning rates (RLRS) using a straightforward and scalable local search algorithm. While grid search could be employed for hyperparameter optimization, it demands precise boundary definitions and entails an exponential number of training runs, making it computationally expensive. Instead, we intentionally adopt a simple local search method to demonstrate that our approach delivers significant improvements even with basic hyperparameter tuning. This approach highlights the robustness of our method and leaves room for future exploration of more sophisticated optimization techniques.
    
This algorithm iteratively adjusts each hyperparameter by scaling its value with factors from a predefined set. If the adjustment improves performance, the change is retained, and the process repeats until no further improvements are observed. In our experiments, we optimized weight decay and initialization scale along with all RLRS values, ensuring a fair comparison. For the baseline, the same method was used to optimize the learning rate at the start and end of the cosine schedule, in addition to weight decay and initialization scale. Further details and the exact algorithm can be found in Appendix.

\subsection{Tuning Small Models}

\begin{figure}
    \centering
    \begin{minipage}[t]{0.335\textwidth}
        \includegraphics[width=\linewidth]{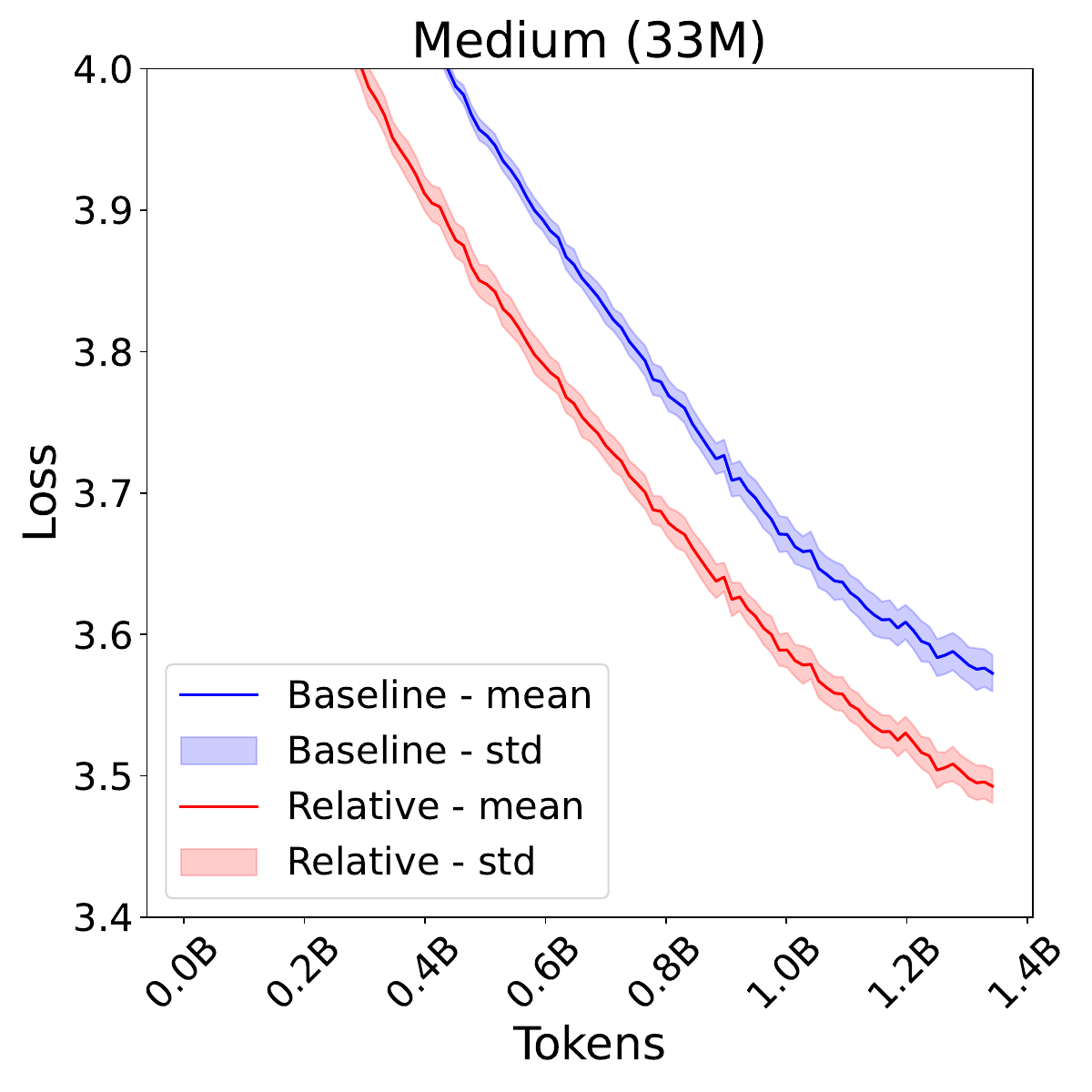}
        \caption{Training loss for Baseline versus RLRS for \moe{33}. }
        \label{fig:loss}
    \end{minipage}
    \hfill
    \centering
    \begin{minipage}[t]{0.3\textwidth}
        \includegraphics[width=\linewidth]{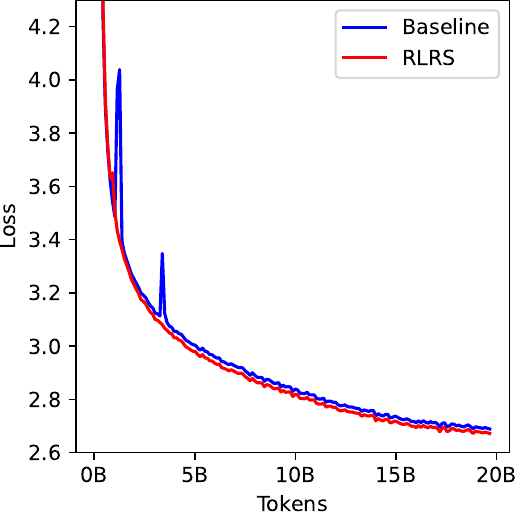}
        \caption{Stabilizing training with RLRS vs. baseline LR for \moe{906}.}
        \label{fig:stable}
    \end{minipage}
    \hfill
    \begin{minipage}[t]{0.3\textwidth}
        \includegraphics[width=\linewidth]{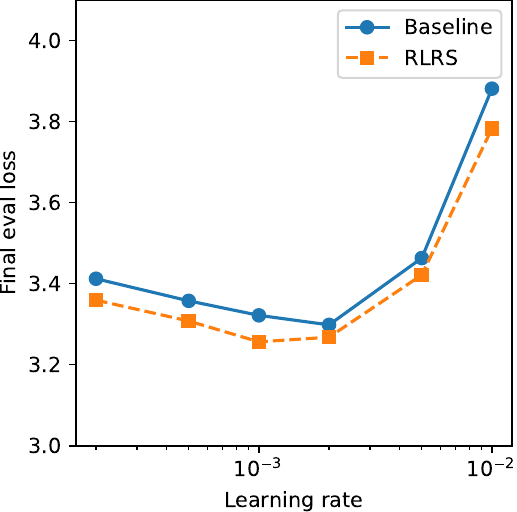}
        \caption{Loss of RLRS and baseline for different  $\eta_{\textit{base}}$ on 
        \moe{113} models. RLRS results in better loss across a range of learning rates.
        }
        \label{fig:sensitivity}
    \end{minipage}
    \label{fig:stability}
\end{figure}

We begin by presenting results obtained with relative learning rates on a small model. The compact size enables a broader search of hyperparameters; therefore, we propose performing the local search on a small model of choice. The details are given in Sec.~\ref{app:local_search} of the Appendix. The relative learning rate values optimized on this smaller model then serve as reference points for extrapolation to larger models.
Adjusting relative rates according to the proposed schedule results in significantly higher sample efficiency, which reduces the training time by up to $23\%$ for MoE and by up to $17\%$ for a dense Transformer (see Table~\ref{tab:small}). The exact relative learning rate values can be found in Section~\ref{sec:analysis}.

\begin{table}[h]
    \centering
\begin{tabular}{ccccccc}
\toprule
Type & LR Type & Base LR & Train Tokens &  Total Params & Speedup \\
\midrule
\dense{34} & baseline & $2 \times 10^{-3}$ & 1.3B & 34M & - \\
& relative & $2 \times 10^{-3}$ & 1.3B & 34M & \roundto{1.17162471395881}{3} \\ 
\midrule
\moe{34} & baseline & $3 \times 10^{-3}$ & 1.3B & 210M & - \\
& relative & $3 \times 10^{-3}$ & 1.3B & 210M & \roundto{1.2278177458033572}{3} \\ 
\bottomrule
\end{tabular}
\vspace{0.1cm}

    \caption{Using RLRS results in faster model convergence. 
    }
    \label{tab:small}
\end{table}

\vspace{-0.3cm}
\subsection{Extrapolation}
\label{sec:exper_extra}



In this section, we show the performance of large models trained with relative learning rate values tuned on small models. We demonstrate that, remarkably, this method improves the training speed of the large models as well.

We use models with $113$M and $906$M active parameters (in the case of MoE, $708$M and $5.67$B total parameters, respectively). The models are trained both in a compute-optimal and overtrained setting.  
As shown in Table~\ref{tab:extrapol}, extrapolating the relative rates results in up to $19\%$ faster training both in the case of MoE and dense model. Furthermore, the improvement is also noticeable in an overtrained MoE model with a token-to-active-parameter ratio that is almost 130. The most far-reaching extrapolations are performed on training runs more than 200$\times$ costly in terms of FLOPs as well as on models 27$\times$ bigger in terms of parameters than the training runs employed in the small-scale relative learning rates tuning. Even in these scaled-up scenarios, RLRS retains noticeably higher sample efficiency.

In Figure~\ref{fig:uplots_base}, we demonstrate that without fine-tuning on a large model, the transferred relative learning rates are still noticeably better than the baseline and generally close to the optimal value. 



\begin{table}[H]
    \centering
\begin{tabular}{ccccccc}
\toprule



Type & LR Type & Base LR & Train Tokens & Total Parameters &  Speedup \\ \midrule

\dense{113} & baseline & $1 \times 10^{-3}$& 2.5B & 113M & - \\
& relative & $1 \times 10^{-3}$ & 2.5B & 113M & \roundto{1.1898734177215189}{3} \\
\midrule
\moe{113} & baseline & $2 \times 10^{-3}$ & 2.5B & 708M & - \\
& relative & $1 \times 10^{-3}$ & 2.5B & 708M & \roundto{1.1898734177215189}{3} \\
\moe{113} & baseline & $1 \times 10^{-3}$ & 14B & 708M & - \\
(overtrained) & relative & $1 \times 10^{-3}$ & 14B & 708M & \roundto{1.1458333333333333}{3} \\
\midrule
\dense{906} & baseline & $5 \times 10^{-4}$ & 20B & 906M & - \\
& relative & $5 \times 10^{-4}$ & 20B & 906M & \roundto{1.0869565217391304}{3} \\
\midrule

\moe{906} & baseline & $2 \times 10^{-4}$ & 20B & 5.67B & - \\
& relative & $2 \times 10^{-4}$ & 20B & 5.67B & \roundto{1.1363636363636365}{3} \\
\bottomrule
\vspace{0.1cm}
\end{tabular}
    \caption{Speedup achieved when extrapolating relative learning rates, RLRS, to larger models.}
\label{tab:extrapol}
\end{table}










\section{Analysis}

\label{sec:analysis}

\subsection{Interpreting Relative Learning Rates}

In this section, we present the numerical results and trends for relative learning rates (RLRS) and analyze them with respect to each layer module. We prioritize MoE models, as RLRS yields more pronounced improvement in this setting.  Interpretations provided below are an attempt to convey intuitions related to why RLRS works well that may be practical to other researchers and are not backed by experimental evidence. Although the values have been determined experimentally, they are often interpretable and aligned to counteract the existing issues of each component. While some of these ideas were also the motivation for this work and could be potentially used to guide the search of the optimal relative learning rates, we suspect that utilizing an exhaustive search will be a more reliable method in the long term.

\paragraph{Embedding.} The relative learning rate $\lambda_{\textit{start}}$ starts high at $5$ and decays to $0.6$. This aggressive early training helps the Embedding stabilize quickly, as it influences the entire network. Later in the training process, the learning rate is reduced to prevent drastic changes in the Embeddings, ensuring the rest of the model can adjust accordingly. As seen in Section~\ref{sec:ablations}, this is the only layer that prefers adjustment of relative learning rate when increasing model size; that is, while other relative learning rates transfer without change, the Embedding's rate should be increased, but only at the start.

\paragraph{Unembedding.} 
Unembedding handles the conversion of the model output into a probability distribution over the tokens in its vocabulary.
We observe that, similarly to the Embedding, the relative learning rate gradually decreases toward the end of training. This behavior also aligns with observations in the literature that weights in the Unembedding may diverge, potentially causing instabilities later in the training process~\citep{chowdhery2022palm, zoph2022st}, which would require reducing gradient values. 

\paragraph{Router + Experts.} Router (or gating network) plays a crucial role in determining which Expert networks are trained during the learning process. ~\citet{xue2024openmoe} observed that the model often learns its routing decisions early in the pre-training phase, and these decisions remain largely fixed throughout training. Once a token is assigned to an Expert, it is rarely reassigned, making it difficult for the model to adapt to new or unseen data during later stages of training. Moreover, there have been conflicting guidelines in literature about suiting the learning rate for MoE models in comparison to their dense counterparts i.e. \citep{zoph2022st} argues that MoE benefit from higher learning rates. On the other hand, multiple studies analyze the instability of MoEs \citep{fedus2022switch}, and the most straightforward approach to alleviate instabilities is to lower the learning rate for the whole model \citep{wortsman2023small}. We suppose that all of these aforementioned analysis are plausible and introduce many contradictory trade-offs that are hard to resolve under the assumption of a fixed learning rate for all modules. The gain from RLRS might come from alleviating these issues. Lowering the learning rate only at the beginning of the training (0.6) for the \textbf{Router} may mitigate instabilities and loss spikes caused by MoE, simultaneously delaying the early router stabilization. Increasing the relative learning rate at the end to 1, allow the model to benefit from the higher value while those issues are no longer present. Similarly, the relative learning rate of an \textbf{Experts} layer starts from the smallest value of $0.3$ to aid stability when the Router is essentially random and prevent early expert specialization, which causes the router to freeze prematurely. It then increases to the highest relative value at the end (1.125), allowing the Experts to fine-tune and benefit from a high learning rate while the Router remains largely fixed.

\paragraph{Attention.} In the Attention layers of the MoE model, the relative learning rate remains unchanged, making it unique in not benefiting from relative rates.

We summarize the Decoupled Learning Rate for both dense and MoE models in Table~\ref{tab:moe_results}.

\begin{table}[h]
    \centering
\begin{tabular}{cccccc}
\toprule  & \text{Embedding} & \text{Unembedding} & \text{Router} & \text{Experts} & \text{Attention} \\
\midrule
\text{start} & $5$ & $0.6$ & $0.6$ & $0.3$ & $1$ \\
\midrule
\text{end} & $0.6$ & $0.4$ & $1$ & $1.125$ &  $1$  \\
\bottomrule
\end{tabular}
\vspace{0.1cm}
    \caption{Relative learning rate values ($\lambda$) for MoE.}
    \label{tab:moe_results}
\end{table}
\begin{table}[h]
    \centering
\begin{tabular}{ccccc}
\toprule  & \text{Embedding} & \text{Unembedding} & \text{Feed-Forward} & \text{Attention} \\
\midrule
\text{start} & $5$ & $1$ & $1$ & $1$ \\
\midrule
\text{end} & $0.6$ & $0.4$ & $0.6$ &  $0.2$  \\
\bottomrule
\end{tabular}
\vspace{0.1cm}
    \caption{Relative learning rate values ($\lambda$) for dense models.}
    \label{tab:dense_results}
\end{table}



\subsection{Stability}

As seen in Figure~\ref{fig:stable}, the baseline exhibits loss spikes that were absent with the relative schedules. This is also intuitive, as MoE models are considered unstable, thus requiring lower learning rates for optimal learning, which, however, affects the speed of training. In our method, the learning rates for both the Router and the Experts start off relatively lower, while they are higher for other parts of the model, resulting in both better stability and convergence.

Large Transformer-based models frequently encounter instabilities, even when using hyperparameters that worked well for smaller models. \citet{wortsman2023small} demonstrate that instabilities in small models with a higher than optimal learning rate can be a good proxy measure for instabilities on a larger scale. Following that, we provide Figure~\ref{fig:sensitivity} comparing the learning rate sensitivity of RLRS and the baseline. We can see that training with relative learning rates outperforms the baseline across various learning rates.
\subsection{Ablations}
\label{sec:ablations}

\begin{figure}[H]  
  \centering
  \includegraphics[width=\textwidth, trim=0 0 0 0, clip]{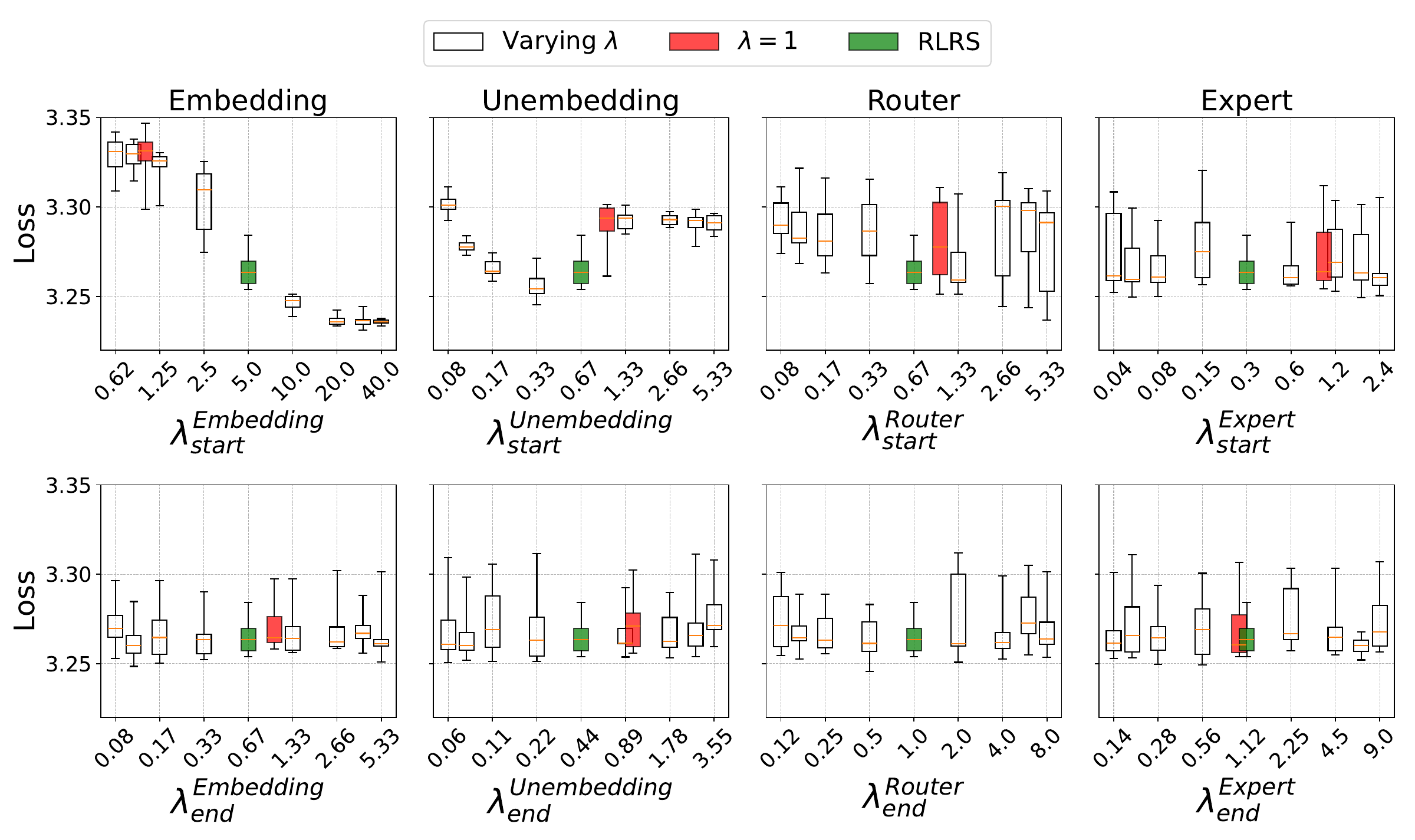}
  \caption{The performance of relative learning rates, $\lambda_{\text{small}}$ extrapolated to larger models. We vary relative values for a given component while keeping all others fixed at their optimal values. The top row explores varying starting learning rates, bottom corresponds to end rates. The optimal $\lambda_{\text{small}}$ (green) is compared to larger and smaller relative rates (white) and the baseline learning rate (red) at $\lambda=1$. The plot shows standard boxplots with 1.5 interquartile range.  Each boxplot is based on 10 experiments on \moe{113}.}
  \label{fig:uplots_base}
\end{figure}

We have shown that relative values tuned on a small model improve the training efficiency of their scaled-up counterparts.
However, the question remains: how effective is this transfer, and can we obtain better results by further tuning relative values on the large model of choice? In Figure~\ref{fig:uplots_base}, we ablate each relative learning rate, $\lambda_{\textit{start}}^m$ and $\lambda_{\textit{end}}^m$ and show the importance of tuning the relative learning rates for individual modules. We conduct tests by evaluating the relative learning rate, transferred from the small model \moe{34} to  \moe{113}, with comparisons to its neighboring values. If the relative rate is not $1$, we also include it as a baseline, accounting for cases where the value is not modified by any relative factor. It is important to note that the improvement brought by the method seems to largely come from the interactions between the relative rates for all the components rather than any specific module. 

While in most cases the transferred rates $\lambda_{\text{start}}^m$ and $\lambda_{\text{end}}^m$ perform consistently well compared to the surrounding values, an interesting exception is the Embedding layer, which shows a clear preference to increase its relative learning rate when increasing the model size. This aligns with \citet{yang2022tensor}, which studies models with increasing width and finds that Embedding is the only layer type whose learning rate should not be scaled down when increasing the model size. This only applies to the relative learning rate at the start, consistent with the Tensor Programs theory, which analyses the stability of the first iteration of the training process.

\section{Related and Future Work}

The literature on learning rates in Machine Learning, particularly for Transformers, highlights the importance of adaptive learning rate schedules. Stochastic Weight Averaging (SWA)~\citep{izmailov2018averaging} utilizes a modified learning rate schedule that applies a decaying learning rate during the initial phase of training, followed by a constant rate for the remainder. 
In \citet{sun2019fine}, the authors introduce layer-wise learning rate decay, which applies higher learning rates to top layers and lower rates to bottom layers. A related concept, discriminative fine-tuning, is discussed in \citet{howard2018universal}. Additionally, \citet{everett2024scaling} explores how various parameterizations and optimizers impact the learning rate transfer in large-scale models and proposes a per-layer (depth-wise) learning rate strategy.

\subsection{Relation to Tensor Programs}
\label{future:tensor_programs}

Our method explores the transfer of relative learning rates; however, the base learning rate must still be independently tuned for the extrapolated model. Approaches such as Tensor Programs~\citep{yang2020tensor, yang2022tensor, everett2024scaling} propose parameterizations that facilitate the transfer of the base learning rate. By combining these two approaches, it may be possible to achieve a zero-shot transfer of RLRS. 

While our methods share similarities with Tensor Programs and draw inspiration from them, our project has a distinct goal. We aim to identify implicit assumptions in the tuning process and decouple parameters to devise a scheme that enables Large Language Models (LLMs) to converge faster. Our extrapolations demonstrate that our optimization scheme depends on the architecture rather than the model size. This scheme is defined relative to the base learning rate, which must be tuned individually for each model size. Our method does not aim to facilitate learning rate transfer between different model sizes and is supported by experimental evidence. We do not mathematically examine the limits of parameter updates in a gradient descent step. A key difference is that our relative values change dynamically during training, and our goal is to enable the model to focus on different modules during pretraining.

\subsection{Fine-Tuning}
Fine-tuning allows users to adapt pre-trained Large Language Models (LLMs) to more specialized tasks. In traditional fine-tuning, certain model components are often ``frozen'' (effectively setting their relative learning rates to zero) to preserve learned knowledge while adapting other parts. Our proposed method introduces a more flexible approach, serving as a continuous alternative to freezing parameters. This enables fine-grained control over information transfer within specific components of the model. Consequently, our method could be particularly applicable to fine-tuning scenarios and complement existing methods that involve freezing parameters. Parameter-Efficient Fine-Tuning (PEFT) techniques, such as LoRA~\citep{hu2021lora}, address this by updating only a subset of parameters while freezing the rest. Our work aligns with more advanced methods like LoRA+~\citep{hayou2024lora+}, which select different learning rates for the adapter matrices, and AdaLoRA~\citep{zhang2023adalora}, which adapts the rank of the LoRA matrices, providing enhanced flexibility in the fine-tuning process.

\section{Conclusion}

We have presented a method for decoupling learning rate schedules across different neural network components, removing the implicit assumption of homogeneity among them. We achieve higher training speed through increased sample efficiency along with greater stability when using RLRS.
Our method applies to any Transformer-based model and significantly enhances performance in Mixture of Experts (MoE) models. By tuning relative learning rates on smaller models, this approach can be used to economically achieve significant improvements in the training of order-of-magnitude larger models. 

\section*{Acknowledgments}

We gratefully acknowledge the Polish high-performance computing infrastructure PLGrid (HPC Center: ACK Cyfronet AGH) for providing computer facilities and support within computational grant no. PLG/2024/017060. This research was partially supported by the ERC PoC Grant EXALT no. 101082299, the National Science Centre (NCN) Grant no. 2020/37/B/ST6/04179, the National Science Centre (NCN) Preludium Grant no. 2022/45/N/ST6/02222, the "European Lighthouse of AI for Sustainability" - ELIAS grant no. 101120237, and the NCBiR grant POIR.01.01.01-00-0433/20. Part of the experiments utilized computational resources provided by \href{https://writer.com/}{Writer}.

\clearpage

\bibliography{iclr2025_conference}

\begin{thebibliography}{22}
\providecommand{\natexlab}[1]{#1}
\providecommand{\url}[1]{\texttt{#1}}
\expandafter\ifx\csname urlstyle\endcsname\relax
  \providecommand{\doi}[1]{doi: #1}\else
  \providecommand{\doi}{doi: \begingroup \urlstyle{rm}\Url}\fi

\bibitem[Chowdhery et~al.(2022)Chowdhery, Narang, Devlin, Bosma, Mishra, Roberts, Barham, Chung, Sutton, Gehrmann, Schuh, Shi, Tsvyashchenko, Maynez, Rao, Barnes, Tay, Shazeer, Prabhakaran, Reif, Du, Hutchinson, Pope, Bradbury, Austin, Isard, Gur-Ari, Yin, Duke, Levskaya, Ghemawat, Dev, Michalewski, Garcia, Misra, Robinson, Fedus, Zhou, Ippolito, Luan, Lim, Zoph, Spiridonov, Sepassi, Dohan, Agrawal, Omernick, Dai, Pillai, Pellat, Lewkowycz, Moreira, Child, Polozov, Lee, Zhou, Wang, Saeta, Diaz, Firat, Catasta, Wei, Meier-Hellstern, Eck, Dean, Petrov, and Fiedel]{chowdhery2022palm}
Aakanksha Chowdhery, Sharan Narang, Jacob Devlin, Maarten Bosma, Gaurav Mishra, Adam Roberts, Paul Barham, Hyung~Won Chung, Charles Sutton, Sebastian Gehrmann, Parker Schuh, Kensen Shi, Sasha Tsvyashchenko, Joshua Maynez, Abhishek Rao, Parker Barnes, Yi~Tay, Noam Shazeer, Vinodkumar Prabhakaran, Emily Reif, Nan Du, Ben Hutchinson, Reiner Pope, James Bradbury, Jacob Austin, Michael Isard, Guy Gur-Ari, Pengcheng Yin, Toju Duke, Anselm Levskaya, Sanjay Ghemawat, Sunipa Dev, Henryk Michalewski, Xavier Garcia, Vedant Misra, Kevin Robinson, Liam Fedus, Denny Zhou, Daphne Ippolito, David Luan, Hyeontaek Lim, Barret Zoph, Alexander Spiridonov, Ryan Sepassi, David Dohan, Shivani Agrawal, Mark Omernick, Andrew~M. Dai, Thanumalayan~Sankaranarayana Pillai, Marie Pellat, Aitor Lewkowycz, Erica Moreira, Rewon Child, Oleksandr Polozov, Katherine Lee, Zongwei Zhou, Xuezhi Wang, Brennan Saeta, Mark Diaz, Orhan Firat, Michele Catasta, Jason Wei, Kathy Meier-Hellstern, Douglas Eck, Jeff Dean, Slav Petrov, and Noah Fiedel.
\newblock Palm: Scaling language modeling with pathways, 2022.

\bibitem[Everett et~al.(2024)Everett, Xiao, Wortsman, Alemi, Novak, Liu, Gur, Sohl-Dickstein, Kaelbling, Lee, and Pennington]{everett2024scaling}
Katie Everett, Lechao Xiao, Mitchell Wortsman, Alexander~A. Alemi, Roman Novak, Peter~J. Liu, Izzeddin Gur, Jascha Sohl-Dickstein, Leslie~Pack Kaelbling, Jaehoon Lee, and Jeffrey Pennington.
\newblock Scaling exponents across parameterizations and optimizers.
\newblock \emph{arXiv preprint arXiv:2407.05872}, 2024.

\bibitem[Fedus et~al.(2022)Fedus, Zoph, and Shazeer]{fedus2022switch}
William Fedus, Barret Zoph, and Noam Shazeer.
\newblock Switch transformers: Scaling to trillion parameter models with simple and efficient sparsity.
\newblock \emph{The Journal of Machine Learning Research}, 23\penalty0 (1):\penalty0 5232--5270, 2022.

\bibitem[Hayou et~al.(2024)Hayou, Ghosh, and Yu]{hayou2024lora+}
Soufiane Hayou, Nikhil Ghosh, and Bin Yu.
\newblock Lora+: Efficient low rank adaptation of large models.
\newblock \emph{arXiv preprint arXiv:2402.12354}, 2024.

\bibitem[Hoffmann et~al.(2024)Hoffmann, Borgeaud, Mensch, Buchatskaya, Cai, Rutherford, de~Las~Casas, Hendricks, Welbl, Clark, Hennigan, Noland, Millican, van~den Driessche, Damoc, Guy, Osindero, Simonyan, Elsen, Vinyals, Rae, and Sifre]{hoffmann2022training}
Jordan Hoffmann, Sebastian Borgeaud, Arthur Mensch, Elena Buchatskaya, Trevor Cai, Eliza Rutherford, Diego de~Las~Casas, Lisa~Anne Hendricks, Johannes Welbl, Aidan Clark, Tom Hennigan, Eric Noland, Katie Millican, George van~den Driessche, Bogdan Damoc, Aurelia Guy, Simon Osindero, Karen Simonyan, Erich Elsen, Oriol Vinyals, Jack~W. Rae, and Laurent Sifre.
\newblock Training compute-optimal large language models.
\newblock In \emph{Proceedings of the 36th International Conference on Neural Information Processing Systems}, NeurIPS '22, 2024.

\bibitem[Howard \& Ruder(2018)Howard and Ruder]{howard2018universal}
Jeremy Howard and Sebastian Ruder.
\newblock Universal language model fine-tuning for text classification.
\newblock In \emph{Proceedings of the 56th Annual Meeting of the Association for Computational Linguistics (Long Papers)}, 2018.

\bibitem[Hu et~al.(2021)Hu, Shen, Wallis, Allen-Zhu, Li, Wang, Wang, and Chen]{hu2021lora}
Edward~J Hu, Yelong Shen, Phillip Wallis, Zeyuan Allen-Zhu, Yuanzhi Li, Shean Wang, Lu~Wang, and Weizhu Chen.
\newblock Lora: Low-rank adaptation of large language models.
\newblock \emph{arXiv preprint arXiv:2106.09685}, 2021.

\bibitem[Izmailov et~al.(2018)Izmailov, Podoprikhin, Garipov, Vetrov, and Wilson]{izmailov2018averaging}
Pavel Izmailov, Dmitrii Podoprikhin, Timur Garipov, Dmitry Vetrov, and Andrew~Gordon Wilson.
\newblock Averaging weights leads to wider optima and better generalization.
\newblock \emph{arXiv preprint arXiv:1803.05407}, 2018.

\bibitem[Loshchilov \& Hutter(2016)Loshchilov and Hutter]{loshchilov2016sgdr}
Ilya Loshchilov and Frank Hutter.
\newblock Sgdr: Stochastic gradient descent with warm restarts.
\newblock \emph{ICLR 2017 (5th International Conference on Learning Representations)}, 2016.

\bibitem[Loshchilov \& Hutter(2019)Loshchilov and Hutter]{loshchilov2019decoupled}
Ilya Loshchilov and Frank Hutter.
\newblock Decoupled weight decay regularization, 2019.

\bibitem[Ludziejewski et~al.(2024)Ludziejewski, Krajewski, Adamczewski, Pióro, Krutul, Antoniak, Ciebiera, Król, Odrzygóźdź, Sankowski, Cygan, and Jaszczur]{ludziejewskiscaling}
Jan Ludziejewski, Jakub Krajewski, Kamil Adamczewski, Maciej Pióro, Michał Krutul, Szymon Antoniak, Kamil Ciebiera, Krystian Król, Tomasz Odrzygóźdź, Piotr Sankowski, Marek Cygan, and Sebastian Jaszczur.
\newblock Scaling laws for fine-grained mixture of experts.
\newblock In \emph{Forty-first International Conference on Machine Learning}, 2024.

\bibitem[Radford et~al.(2018)Radford, Narasimhan, Salimans, and Sutskever]{radford2018improving}
Alec Radford, Karthik Narasimhan, Tim Salimans, and Ilya Sutskever.
\newblock Improving language understanding by generative pre-training.
\newblock 2018.

\bibitem[Raffel et~al.(2019)Raffel, Shazeer, Roberts, Lee, Narang, Matena, Zhou, Li, and Liu]{raffel2023exploring}
Colin Raffel, Noam~M. Shazeer, Adam Roberts, Katherine Lee, Sharan Narang, Michael Matena, Yanqi Zhou, Wei Li, and Peter~J. Liu.
\newblock Exploring the limits of transfer learning with a unified text-to-text transformer.
\newblock \emph{J. Mach. Learn. Res.}, 21:\penalty0 140:1--140:67, 2019.

\bibitem[Rajbhandari et~al.(2022)Rajbhandari, Li, Yao, Zhang, Aminabadi, Awan, Rasley, and He]{pmlr-v162-rajbhandari22a}
Samyam Rajbhandari, Conglong Li, Zhewei Yao, Minjia Zhang, Reza~Yazdani Aminabadi, Ammar~Ahmad Awan, Jeff Rasley, and Yuxiong He.
\newblock {D}eep{S}peed-{M}o{E}: Advancing mixture-of-experts inference and training to power next-generation {AI} scale.
\newblock In Kamalika Chaudhuri, Stefanie Jegelka, Le~Song, Csaba Szepesvari, Gang Niu, and Sivan Sabato (eds.), \emph{Proceedings of the 39th International Conference on Machine Learning}, volume 162 of \emph{Proceedings of Machine Learning Research}, pp.\  18332--18346. PMLR, 17--23 Jul 2022.
\newblock URL \url{https://proceedings.mlr.press/v162/rajbhandari22a.html}.

\bibitem[Sun et~al.(2019)Sun, Qiu, Xu, and Huang]{sun2019fine}
Chi Sun, Xipeng Qiu, Yige Xu, and Xuanjing Huang.
\newblock How to fine-tune bert for text classification?
\newblock In \emph{Chinese computational linguistics: 18th China national conference, CCL 2019, Kunming, China, October 18--20, 2019, proceedings 18}, pp.\  194--206. Springer, 2019.

\bibitem[Touvron et~al.(2023)Touvron, Lavril, Izacard, Martinet, Lachaux, Lacroix, Rozière, Goyal, Hambro, Azhar, Rodriguez, Joulin, Grave, and Lample]{touvron2023llama}
Hugo Touvron, Thibaut Lavril, Gautier Izacard, Xavier Martinet, Marie-Anne Lachaux, Timothée Lacroix, Baptiste Rozière, Naman Goyal, Eric Hambro, Faisal Azhar, Aurelien Rodriguez, Armand Joulin, Edouard Grave, and Guillaume Lample.
\newblock Llama: Open and efficient foundation language models, 2023.

\bibitem[Wortsman et~al.(2023)Wortsman, Liu, Xiao, Everett, Alemi, Adlam, Co-Reyes, Gur, Kumar, Novak, Pennington, Sohl-dickstein, Xu, Lee, Gilmer, and Kornblith]{wortsman2023small}
Mitchell Wortsman, Peter~J. Liu, Lechao Xiao, Katie Everett, Alex Alemi, Ben Adlam, John~D. Co-Reyes, Izzeddin Gur, Abhishek Kumar, Roman Novak, Jeffrey Pennington, Jascha Sohl-dickstein, Kelvin Xu, Jaehoon Lee, Justin Gilmer, and Simon Kornblith.
\newblock Small-scale proxies for large-scale transformer training instabilities.
\newblock \emph{arXiv preprint arXiv:2309.14322}, 2023.

\bibitem[Xue et~al.(2024)Xue, Zheng, Fu, Ni, Zheng, Zhou, and You]{xue2024openmoe}
Fuzhao Xue, Zian Zheng, Yao Fu, Jinjie Ni, Zangwei Zheng, Wangchunshu Zhou, and Yang You.
\newblock Openmoe: An early effort on open mixture-of-experts language models.
\newblock \emph{arXiv preprint arXiv:2402.01739}, 2024.

\bibitem[Yang(2020)]{yang2020tensor}
Greg Yang.
\newblock Tensor programs ii: Neural tangent kernel for any architecture.
\newblock \emph{arXiv preprint arXiv:2006.14548}, 2020.

\bibitem[Yang et~al.(2022)Yang, Hu, Babuschkin, Sidor, Liu, Farhi, Ryder, Pachocki, Chen, and Gao]{yang2022tensor}
Greg Yang, Edward~J Hu, Igor Babuschkin, Szymon Sidor, Xiaodong Liu, David Farhi, Nick Ryder, Jakub Pachocki, Weizhu Chen, and Jianfeng Gao.
\newblock Tensor programs v: Tuning large neural networks via zero-shot hyperparameter transfer.
\newblock \emph{arXiv preprint arXiv:2203.03466}, 2022.

\bibitem[Zhang et~al.(2023)Zhang, Chen, Bukharin, Karampatziakis, He, Cheng, Chen, and Zhao]{zhang2023adalora}
Qingru Zhang, Minshuo Chen, Alexander Bukharin, Nikos Karampatziakis, Pengcheng He, Yu~Cheng, Weizhu Chen, and Tuo Zhao.
\newblock Adalora: Adaptive budget allocation for parameter-efficient fine-tuning.
\newblock In \emph{International Conference on Learning Representations (ICLR)}, 2023.

\bibitem[Zoph et~al.(2022)Zoph, Bello, Kumar, Du, Huang, Dean, Shazeer, and Fedus]{zoph2022st}
Barret Zoph, Irwan Bello, Sameer Kumar, Nan Du, Yanping Huang, Jeff Dean, Noam Shazeer, and William Fedus.
\newblock St-moe: Designing stable and transferable sparse expert models.
\newblock \emph{arXiv preprint arXiv:2202.08906}, 2022.

\end{thebibliography}
\bibliographystyle{iclr2025_conference}

\clearpage
\appendix

\section{Finding Decoupled Relative Learning Rates} \label{app:local_search}


Our method involves determining a set of relative learning rates. While these hyperparameters could be optimized using a straightforward grid search, such a procedure requires carefully setting the search boundaries and involves an exponential number of training runs. In our experiments, we opt for a more scalable local search algorithm, which is described below.

\begin{algorithm}
\caption{Local Search}
\begin{algorithmic}[1]
\State Iterate over the set of hyperparameters.
\State For a given hyperparameter, multiply its value by a factor from $\{\frac{1}{5}, \frac{2}{3}, \frac{3}{2}, \frac{5}{1}\}$
\State Run experiments, and if there is an improvement, adjust the hyperparameter value.
\State If any change has been made among all hyperparameters, return to Step $1$.
\end{algorithmic}
\label{algo_local}
\end{algorithm}

We note that Algorithm~\ref{algo_local} is a relatively simple optimization method, and we expect that a more complex alternative could either converge faster or find an even better set of hyperparameters. To ensure proper configuration, we optimized the weight decay and initialization scale along with all RLRS values. For the baseline, the same algorithm was used to find the learning rate at the start and at the end of the cosine schedule, along with weight decay and initialization scale. 

\newpage
\section{Additional Figures}

\begin{figure}[H]  
  \centering
  \includegraphics[width=\textwidth]{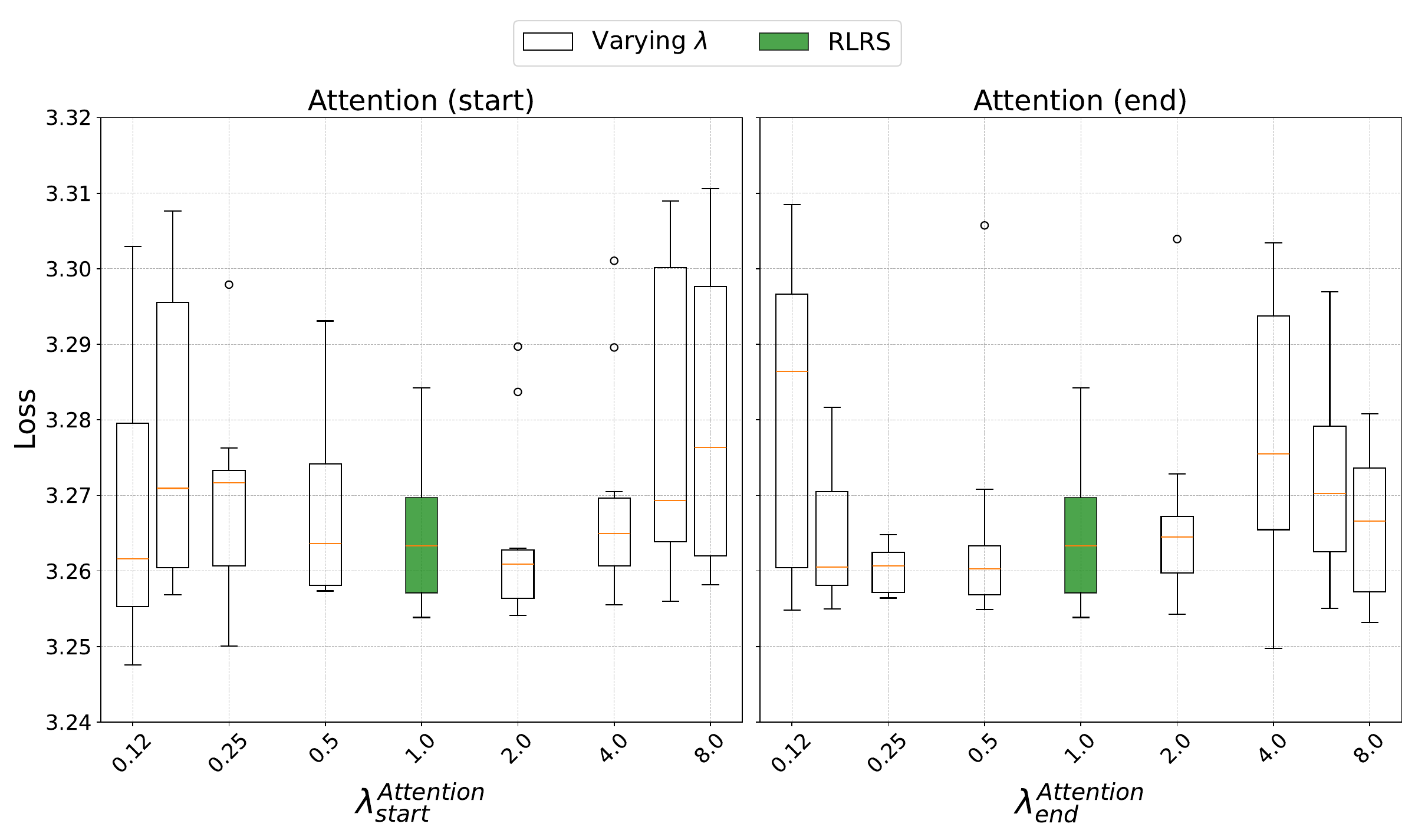} 
  \caption{Varying $\lambda_{\text{start}}^{\text{Attention}}$ and $\lambda_{\text{end}}^{\text{Attention}}$ for \moe{116}. For this component, our optimization algorithm kept the relative value unchanged ($\lambda = 1.0$). }
  \label{fig:uplots_attention}
\end{figure}

\begin{figure}[h]
    \centering
    \begin{minipage}[t]{0.45\textwidth}
        \includegraphics[width=\linewidth]{figs/medium_relative_vs_baseline.pdf}
        \caption{Training loss for Baseline versus RLRS for \moe{33} ($210$M total parameters), averaged among 3 runs.}
        \label{fig:medium_relative_vs_baseline}
    \end{minipage}
    \hfill
    \begin{minipage}[t]{0.45\textwidth}
        \includegraphics[width=\linewidth]{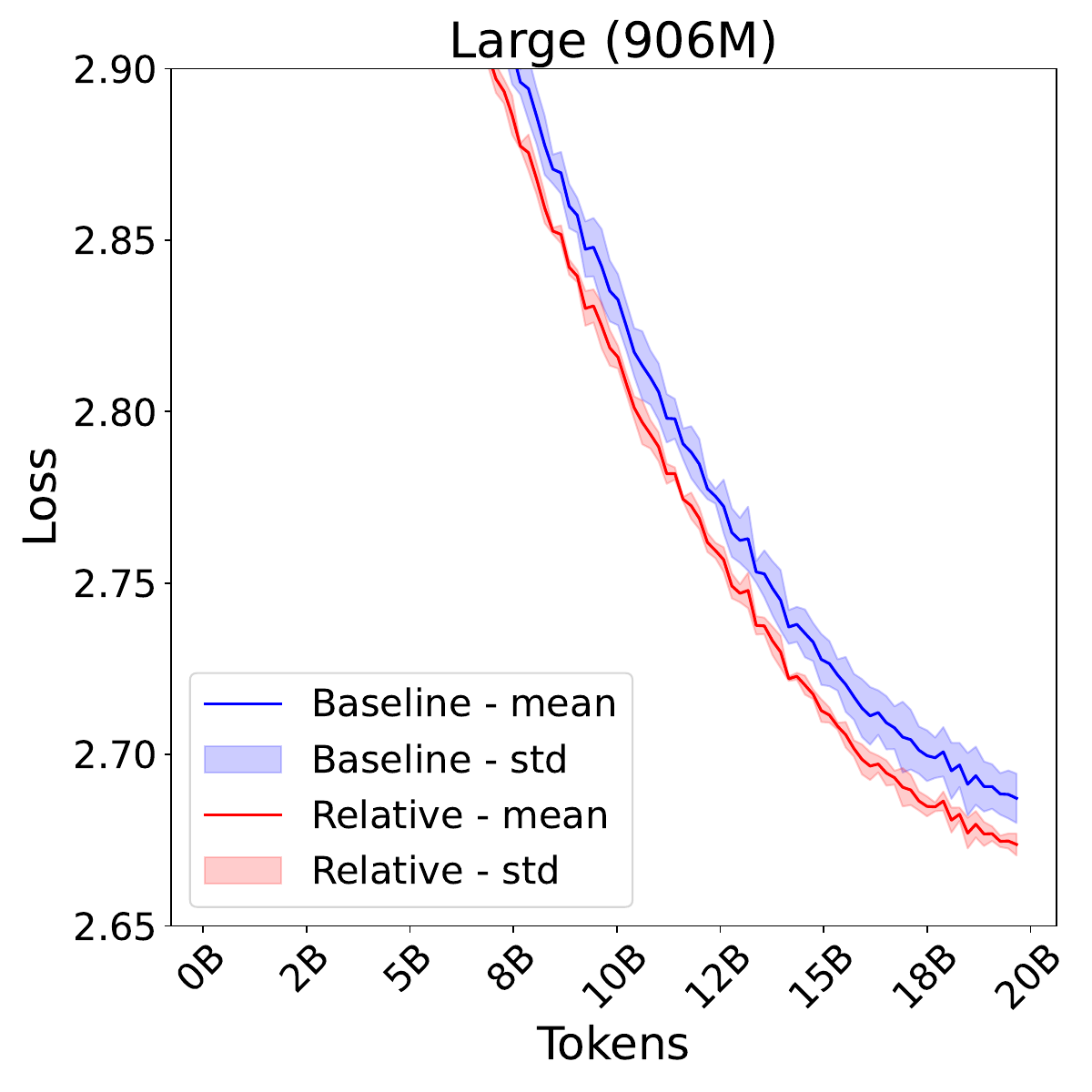}
        \caption{Training loss for Baseline versus RLRS for \moe{906} ($5.67$B total parameters) with hyperparameters extrapolated from \moe{33}, averaged among 3 runs. }
    \end{minipage}
    \label{fig:large_relative_vs_baseline}
\end{figure}

\begin{figure}[h]  
  \centering
  \includegraphics[width=0.9\textwidth]{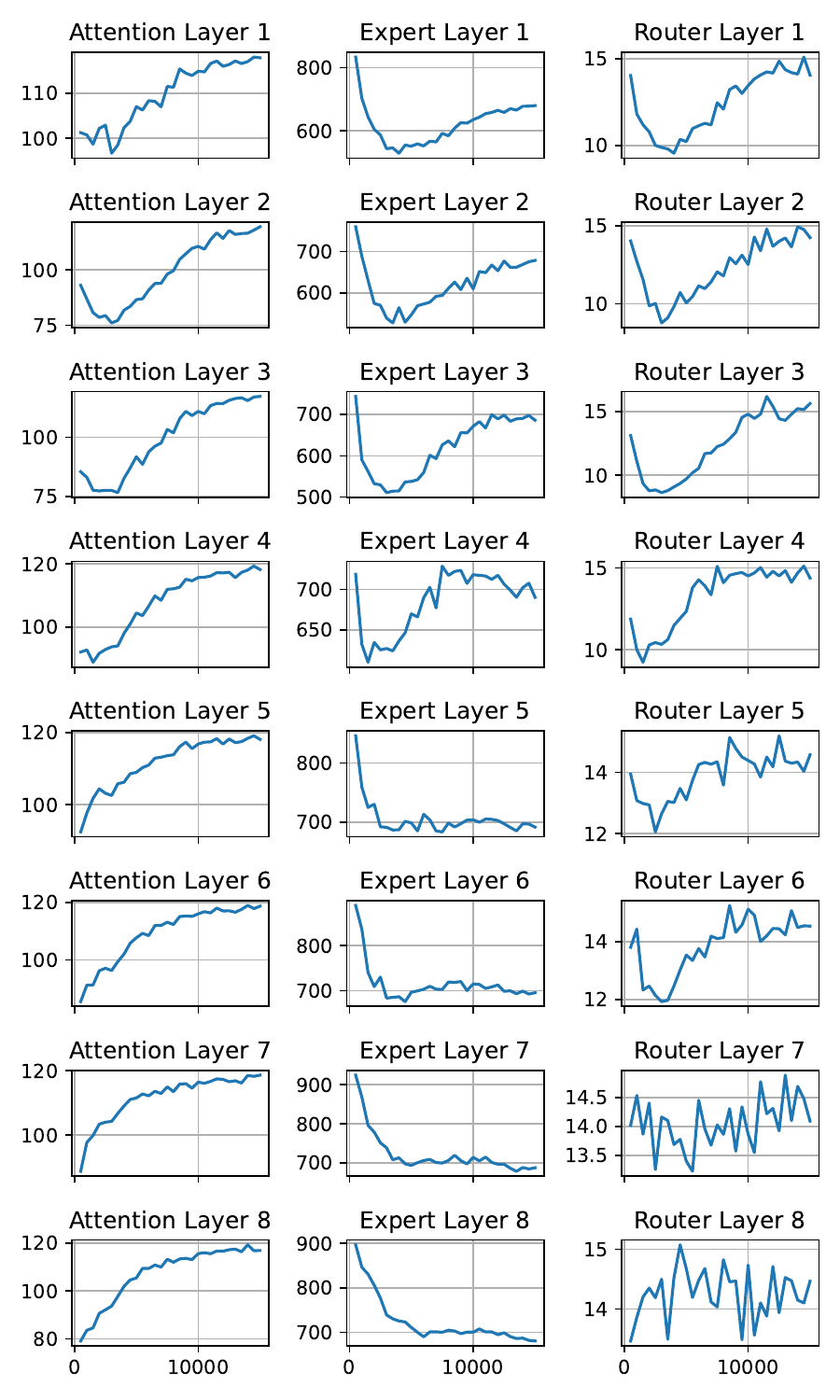} 
  \caption{Magnitudes of weight updates for different components of all 8 layers of \moe{33} trained without RLRS, right before applying learning rate.}
  \label{fig:gradients_all}
\end{figure}



\end{document}